\documentclass[letterpaper, 10 pt, conference]{ieeeconf}
\IEEEoverridecommandlockouts                            
\usepackage{graphics}
\usepackage{times}   
\usepackage{amsmath} 
\usepackage{amssymb} 
\usepackage{graphicx}
\usepackage{subfig}
\usepackage[noline, ruled]{algorithm2e}
\usepackage{booktabs}
\usepackage[noend]{algpseudocode}
\usepackage{subfloat}
\usepackage{xspace}
\usepackage{pifont}
\usepackage{hyperref}
\usepackage{textcomp}
\newcommand{\cmark}{\ding{51}}%
\newcommand{\xmark}{\ding{55}}%
\newcommand{\norm}[1]{\big\lVert#1\big\rVert}

\usepackage[font=small]{caption}

\def\secref#1{Sec.~\ref{#1}}
\def\figref#1{Fig.~\ref{#1}}
\def\tabref#1{Tab.~\ref{#1}}
\def\eqref#1{Eq.~(\ref{#1})}
\def\algref#1{Alg.~\ref{#1}}

\newcommand\etal{\emph{et al.}}

\usepackage{array}
\newcolumntype{L}[1]{>{\raggedright\let\newline\\\arraybackslash\hspace{0pt}}m{#1}}
\newcolumntype{C}[1]{>{\centering\let\newline\\\arraybackslash\hspace{0pt}}m{#1}}
\newcolumntype{R}[1]{>{\raggedleft\let\newline\\\arraybackslash\hspace{0pt}}m{#1}}

\makeatletter
\DeclareRobustCommand\onedot{\futurelet\@let@token\@onedot}
\def\@onedot{\ifx\@let@token.\else.\null\fi\xspace}

\def\eg{e.g\onedot} 
\def\ie{i.e\onedot} 
 
\def\etc{etc\onedot} 
\def\wrt{w.r.t\onedot} 
\def\etal{\emph{et al}\onedot}
\makeatother

\title{\LARGE \bf ReFusion: 3D Reconstruction in Dynamic Environments\\ for RGB-D Cameras Exploiting Residuals}

\author{Emanuele Palazzolo \and Jens Behley \and Philipp Lottes \and Philippe Gigu\`{e}re \and Cyrill Stachniss
  \thanks{Emanuele Palazzolo, Jens Behley, Philipp Lottes and Cyrill Stachniss are with the University of Bonn, Germany. Philippe Gigu\`{e}re is with the Laval University, Qu{\'e}bec, Canada.}%
  \thanks{This work has partly been supported by
  the DFG under the grant number FOR~1505: Mapping on Demand, under the grant number BE 5996/1-1, and under Germany's Excellence Strategy, EXC-2070 - 390732324 (PhenoRob).
  }%
}

\begin{document}
\maketitle
\thispagestyle{empty}
\pagestyle{empty}

\begin{abstract}
Mapping and localization are essential capabilities of robotic systems. Although the majority of mapping systems focus on static environments, the deployment in real-world situations requires them to handle dynamic objects. In this paper, we propose an approach for an RGB-D sensor that is able to consistently map scenes containing multiple dynamic elements. For localization and mapping, we employ an efficient direct tracking on the truncated signed distance function (TSDF) and leverage color information encoded in the TSDF to estimate the pose of the sensor. The TSDF is efficiently represented using voxel hashing, with most computations parallelized on a GPU. For detecting dynamics, we exploit the residuals obtained after an initial registration, together with the explicit modeling of free space in the model. We evaluate our approach on existing datasets, and provide a new dataset showing highly dynamic scenes. These experiments show that our approach often surpass other state-of-the-art dense SLAM methods. We make available our dataset with the ground truth for both the trajectory  of the RGB-D sensor obtained by a motion capture system and the model of the static environment using a high-precision terrestrial laser scanner. Finally, we release our approach as open source code.
\end{abstract}

\section{Introduction}
\label{sec:intro}

\begin{figure}[t]
  \centering
  \includegraphics[width=0.75\linewidth]{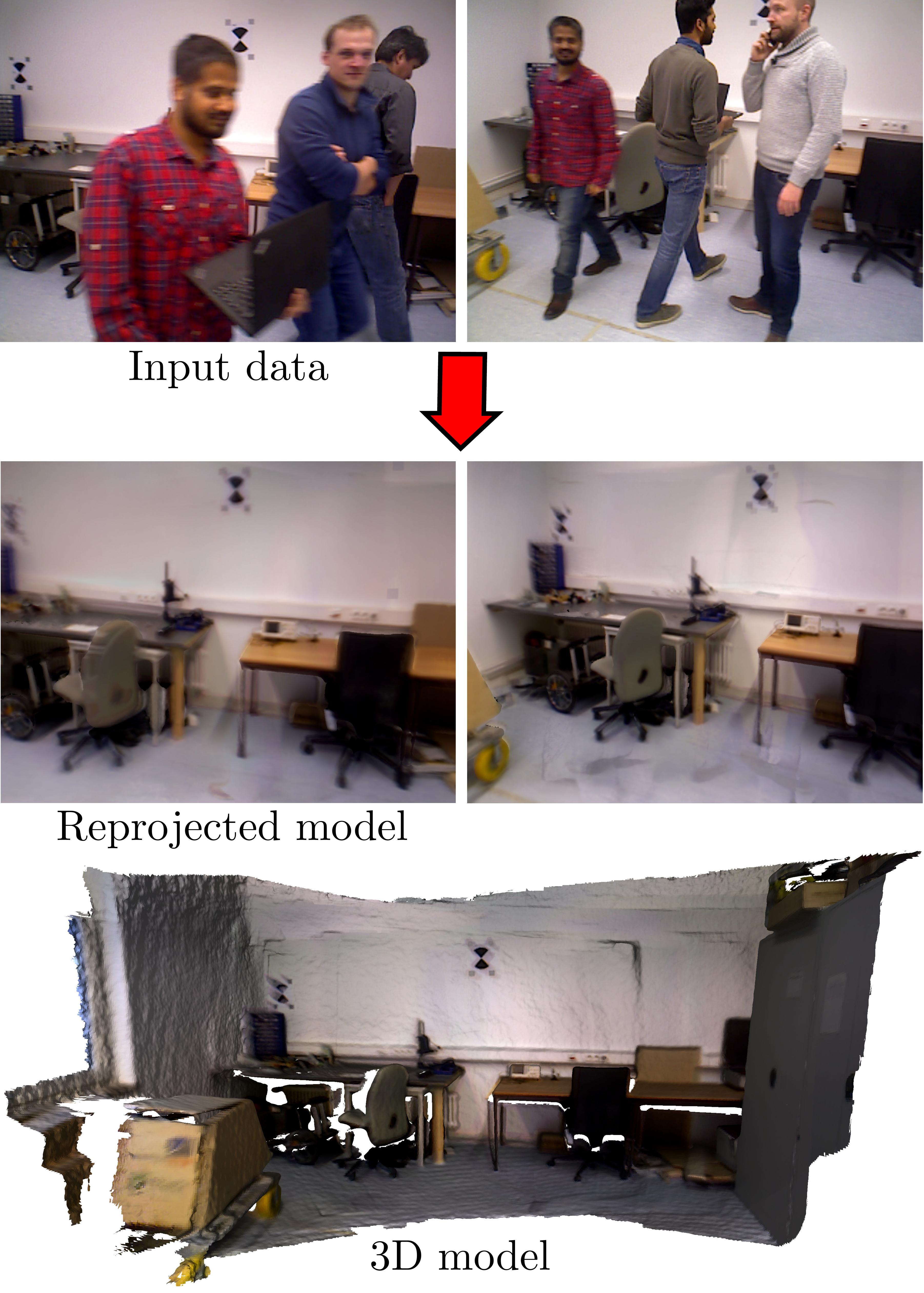}
  \caption{Result of our approach. Top: RGB frames from our dataset containing dynamics. Center: Reconstructed model without dynamics reprojected onto the two frames from above. Bottom: Final mesh  without the dynamics.}
\label{fig:motivation}
\end{figure}

Mapping and localization are essential capabilities of robotic systems operating in real-world environments.
Simultaneous localization and mapping, or SLAM, is usually solved in an alternating fashion, where one determines the pose \wrt the map built so far and then use the estimated pose to update the map. SLAM is especially challenging in dynamic environments, since a robot needs to build a consistent map. This entails estimating simultaneously which parts of the environment are static or moving. In particular, moving objects may cause wrong correspondences, deteriorate the ability to estimate correct poses and hence corrupt the map.

In this paper, we propose \emph{ReFusion}: a novel approach for dense indoor mapping that is robust to dynamic elements. Our approach is completely geometric, and does not rely on an explicit semantic interpretation of the scene. More specifically, we do not employ a deep neural network to detect specific dynamic classes, in contrast to other recent approaches~\cite{ruenz2018ismar,bescos2018ral}. In contrast to other recent purely geometric approaches~\cite{runz2017icra,scona2018icra}, we do not represent the model using surfels, but in the form of a truncated sign distance function (TSDF). This allows our technique to directly generate a mesh of the environment, which can be useful in application such as virtual and augmented reality. Moreover, the TSDF representation can be useful for planning, since it provides, by definition, the distance to the closest obstacle.

The main contribution of this paper is a novel and efficient  SLAM algorithm, based on a TSDF representation, that is robust to dynamics via pure geometric filtering. It is inspired by the work of Canelhas \etal~\cite{canelhas2013iros} combined with voxel hashing~\cite{niessner2013siggraph}. We propose to detect dynamics by exploiting the residuals obtained from the registration, in combination with the explicit representation of free space in the environment. This allows our approach to be class agnostic, \ie, it does not rely on a detector trained on specific categories of dynamic objects. Moreover, the dynamic objects are not explicitly tracked, therefore our approach is not limited by the number of moving objects or by their speed. \figref{fig:motivation} shows two example RGB frames from a dynamic scene and the resulting model built by our approach.

We evaluated ReFusion on the TUM RGB-D dataset~\cite{sturm2012iros}, as well as on our own dataset, showing the versatility and robustness of our approach, reaching in several scenes equal or better performance than other dense SLAM approaches. In addition, we publicly release our dataset, containing $24$ highly-dynamic scenes recorded with an RGB-D sensor, together with ground truth trajectories obtained using a motion capture system. Furthermore,  we provide a ground truth 3D model of the static parts of the environment in the form of a high resolution point cloud acquired with a terrestrial laser scanner. To the best of our knowledge, this is the first dataset containing dynamic scenes that also includes the ground truth model for the static part of the environment. Finally, we publicly share the open source implementation of our approach.

In sum, we make two key claims:
our \mbox{RGB-D} mapping approach (i) is robust to dynamic elements in the environment and provides a camera tracking performance on par or better than state-of-the-art dense SLAM approaches, and
(ii) provides a dense 3D model that contains only the static parts of the environment, which is more accurate than other state-of-the-art approaches when compared to the ground truth model.

\section{Related Work}
\label{sec:related}

With the advent of inexpensive RGB-D cameras, many approaches for mapping using such sensors were proposed \cite{stachniss2016handbook-slamchapter,zollhoefer2018eg}.
The seminal paper of Newcombe \etal~\cite{newcombe2011ismar} showed the prospects of TSDF-based RGB-D mapping by generating accurate, high detailed maps using only depth information. It paved the way for several improvements increasing the versatility and fidelity of RGB-D mapping.
As the approach relies on a fixed voxel grid, volumes that can be mapped are limited. Subsequent approaches explore compression of this grid. For instance, Steinbr\"ucker \etal~\cite{steinbruecker2014icra} use an Octree instead of a voxel grid.
Nie\ss ner \etal~\cite{niessner2013siggraph}, on the other hand, propose to only allocate voxel blocks close to the mapped surface, and address them in constant time via hashing.
K\"ahler \etal~\cite{kaehler2016icra} extend the idea of voxel hashing by using a hierarchy of voxel blocks with different resolutions.
To alleviate the need for raycasting for generating the model image for registration, 
Canelhas \etal~\cite{canelhas2013iros} and Bylow \etal~\cite{bylow2013rss} 
propose to directly exploit the TSDF for evaluation of the residuals and computation of the jacobians
within the error minimization.

Besides using a TSDF, another popular representation of the model are surfels, which are disks with a normal and a radius.
Keller \etal~\cite{keller20133dv} use surfels to represent the model of the environment. Whelan \etal~\cite{whelan2015rss} extend the approach with a deformation graph, which allows for long-range corrections of the map via loop closures.

An alternative approach was proposed by Della Corte \etal~\cite{della-corte2018icra}. Their approach registers two consecutive RGB-D frames directly upon each other by minimizing the photometric error. They integrate multiple cues such as depth, color and normal information in a unified way. Other approaches exploit image sequences~\cite{naseer2015iros,vysotska2016ral}. %

Usually, mapping approaches assume a static environment and therefore handle moving objects mostly through outliers rejection.
By discarding information that disagrees with the current measurements, one can handle implicitly dynamic objects~\cite{newcombe2011ismar}.
Other approaches model the dynamic parts of the environment explicitly~\cite{keller20133dv, scona2018icra} and filter these before the integration into the model.
Keller \etal~\cite{keller20133dv} use outliers in point correspondences during ICP as seed for segmentation of dynamics. Corresponding model surfels inside the segments are then marked as unstable.
In contrast, R\"unz \etal~\cite{runz2017icra} explicitly track moving objects given by a segmentation process using either motion or semantic cues provided by class-agnostic 
object proposals \cite{pinheiro2016eccv}.
Scona \etal~\cite{scona2018icra} extend ElasticFusion~\cite{whelan2015rss} to incorporate only clusters that correspond to the static environment.
Distinguishing static and dynamic parts of the environment is achieved by jointly estimating the camera pose and whether clusters are static or dynamic.
Bescos \etal~\cite{bescos2018ral} combine a geometric approach with a deep learning segmentation to enable removal of dynamics in an ORB-SLAM2 system~\cite{mur-artal2017tro}.
Similarly, R{\"u}nz \etal~\cite{ruenz2018ismar} exploit deep learning segmentation, refined with a geometric approach, to detect objects. In addition, they reconstruct and track  detected objects independently.

In contrast to the aforementioned approaches, we propose a mapping approach robust to dynamics which relies on the residuals from the registration and on the detected free space. In this way, our approach is able to detect any kind of dynamics without relying on specific classes or models, and without explicitly tracking dynamic objects.

\section{Our Approach}
\label{sec:main}

\begin{figure}[t]
  \centering
  \vspace{2mm}
  \includegraphics[width=0.99\linewidth]{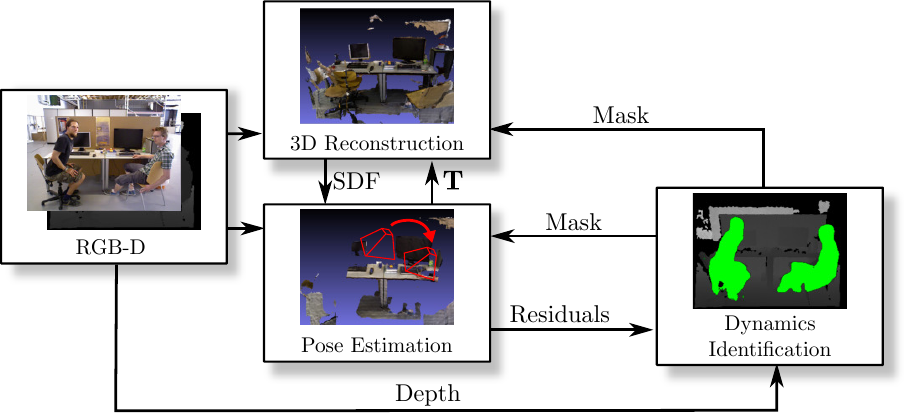}
  \caption{Overview of our approach. Given data from the RGB-D sensor, we first perform an initial pose estimation. Then, we use the obtained residuals, together with the depth information, to identify dynamic parts of the scene. The filtered images are then used to refine the pose $\mathbf{T}$ \wrt the TSDF given by our 3D reconstruction. With the updated sensor pose, we finally integrate the measurements into our 3D reconstruction of the environment.}
\label{fig:concept}
\end{figure}

\figref{fig:concept} illustrates the key processing steps of the proposed approach.
Given the color and depth information of an RGB-D sensor, like Microsoft's Kinect, we first perform an initial pose estimation by exploiting directly the TSDF of our model representation. By observing the residuals obtained from such registration, we detect the dynamic elements in the scene. With the filtered sensor information, where we discard regions containing dynamics, we further refine the pose of the sensor. With this refined estimated pose, we then integrate the sensor measurements, \ie, depth and color, into the model.

\subsection{Model Representation}

In our approach, we represent the model of the environment using a truncated signed distance function (TSDF) as originally proposed by
Curless and Levoy~\cite{curless1996siggraph}. We briefly recap it here for the sake of a self-contained description. The idea is to represent the world
with a 3D voxel grid in which each voxel contains a SDF value. The SDF is a function
$V_\mathrm{SDF}(\mathbf{x}): \mathbb{R}^3 \rightarrow \mathbb{R}$ that returns,
given a point in space, its signed distance to the nearest surface. This
distance is positive if the point is in front of the surface and negative otherwise. In this way, the surface is implicitly represented
by the set of points for which $V_\mathrm{SDF}(\mathbf{x}) = 0$. In practice,
the SDF values are truncated to a given maximum value.
In addition to the TSDF, each voxel contains a weight that represents how reliable the SDF value is at that location, as well as the color obtained by projecting it onto the RGB image. The weight allows us to update each voxel using a running weighted average, leading to an improved robustness to outliers~\cite{curless1996siggraph}. The color information enables both the texturing of the mesh and the use of intensity information during the registration.

In our implementation, we do not preallocate a 3D voxel grid of a fixed size. Instead, we allocate voxel dynamically and index them using a spatial hashing function, similarly to Nie{\ss}ner \etal~\cite{niessner2013siggraph}. Representing the world sparsely by only allocating the needed blocks of voxels enables the reconstruction of larger scenes, while eliminating the need for restricting the maximum size of the scene in advance. Note that we process every voxel in parallel on the GPU, since the voxels are assumed to be mutually independent, leading to a considerable speed-up of the computations.

\subsection{Pose Estimation}\label{sec:pose_registration}

For estimating the pose of the sensor, we use a point-to-implicit approach as proposed independently  by
Canelhas \etal~\cite{canelhas2013iros} and Bylow \etal~\cite{bylow2013rss}. In contrast to KinectFusion~\cite{newcombe2011ismar} and similar methods, our approach does not generate synthetic views from the model. Instead, we use an
alternative technique and directly align an incoming point cloud from the sensor to the SDF, since the SDF provides by definition the distance of a point to the closest
surface, \ie, it is possible to directly use the SDF value as an error function. In addition to the existing point-to-implicit
techniques, we exploit the color information contained in the model to improve the alignment.

Each frame of an RGB-D sensor consists of a depth image and a color image.
Given a pixel $\mathbf{p} = \begin{bmatrix}u & v \end{bmatrix}^\top$, we define
the functions $D(\mathbf{p}): \mathbb{R}^2 \rightarrow \mathbb{R}$ and
$I(\mathbf{p}): \mathbb{R}^2 \rightarrow \mathbb{R}$, which map a pixel 
to its depth and its intensity, respectively.

We denote by $\mathbf{x}$ the 3D point resulting from the back-projection of a pixel
$\mathbf{p}$:
\begin{eqnarray}\label{eq:backprojection}
\mathbf{x} &=& \begin{bmatrix}
\frac{u-c_x}{f_x}D(\mathbf{p}) \\
\frac{v-c_y}{f_y}D(\mathbf{p}) \\
D(\mathbf{p})
\end{bmatrix},
\end{eqnarray}
where $c_x$, $c_y$, $f_x$ and $f_y$ are the intrinsic parameters of the camera, assuming a pinhole camera  model.

We represent a camera pose as a 3D transformation $\mathbf{T}\in\mathbb{SE}(3)$.
A \emph{small} rigid-body motion can be written in minimal form using the Lie algebra representation
$\boldsymbol{\xi}\in\mathfrak{se}_3$, which can be converted into the corresponding rigid transformation $\mathbf{T}' \in \mathbb{SE}(3)$ using the exponential map $\mathbf{T}^\prime = \exp({\hat{\boldsymbol{\xi}}})$, where $\hat{\boldsymbol{\xi}}$ is the corresponding skew symmetric matrix of $\boldsymbol{\xi}$. 

To represent the current model, we use a voxel-based representation, where we store in each voxel the TSDF value and color information. We define the functions $V_\mathrm{SDF}(\mathbf{x})$ and $V_\mathrm{I}(\mathbf{x})$ that return respectively the SDF and the intensity from the model at position $\mathbf{x}$. We obtain the intensity value $I$ from the RGB color information stored in the voxels, \ie, $I = 0.2126 R + 0.7152 G + 0.0722 B$. 
We furthermore use trilinear interpolation on the SDF and intensity of neighboring voxels to alleviate discretization effects of the voxel grid.

As mentioned before, we directly exploit
the TSDF to define the error function, since it directly represents the distance
of a point to the nearest surface of the model. We define the error function relative to the depth as:
\begin{eqnarray}\label{eq:errordepth}
E_d(\hat{\boldsymbol{\xi}}) &=& \sum_{i=1}^N\underbrace{\norm{V_\mathrm{SDF}\big(\exp(\hat{\boldsymbol{\xi}})\mathbf{T}\mathbf{x}_i\big)}^2}_{r_i},
\end{eqnarray}
where $N$ is the number of pixels in the image, and $\mathbf{x}_i, i \in 1,\dots, N$, is
the 3D point corresponding to the $i$-th pixel $\mathbf{p}_i$, computed using \eqref{eq:backprojection}. The value $r_i$ corresponds to the residual for the $i$-th pixel.

We additionally use the color information to improve the alignment. In contrast
to most of the state-of-the-art approaches \cite{whelan2015rss,vempati2017iros}, we do not
render a synthetic view from the model. Instead, we directly operate on the intensity obtained from the color
information stored in the voxels. We define the error function relative
to the intensity as the photometric error between the intensity of the pixels of the
current image and the intensity of the corresponding voxels in the model:
\begin{eqnarray}\label{eq:errorcolor}
E_c(\hat{\boldsymbol{\xi}}) &=& \sum_{i=1}^N\norm{V_\mathrm{I}\big(\exp(\hat{\boldsymbol{\xi}})\mathbf{T}\mathbf{x}_i\big)-I(\mathbf{p}_i)}^2,
\end{eqnarray}
and the joint error function is given by:
\begin{eqnarray}\label{eq:errorjoint}
E(\hat{\boldsymbol{\xi}}) &=& E_d(\hat{\boldsymbol{\xi}}) + w_cE_c(\hat{\boldsymbol{\xi}}),
\end{eqnarray}
with $w_c$ weighting the contribution of the intensity information \wrt the depth information.

We solve this least-squares problem using Levenberg-Marquardt on three different coarse-to-fine sub-sampling of the input images to speed-up convergence.
We furthermore exploit the GPU by processing each pixel in parallel to further accelerate the minimization process.

\subsection{Dynamics Detection}

\begin{figure}[t]
  \centering
  \subfloat[RGB Image]{\includegraphics[width=0.48\columnwidth]{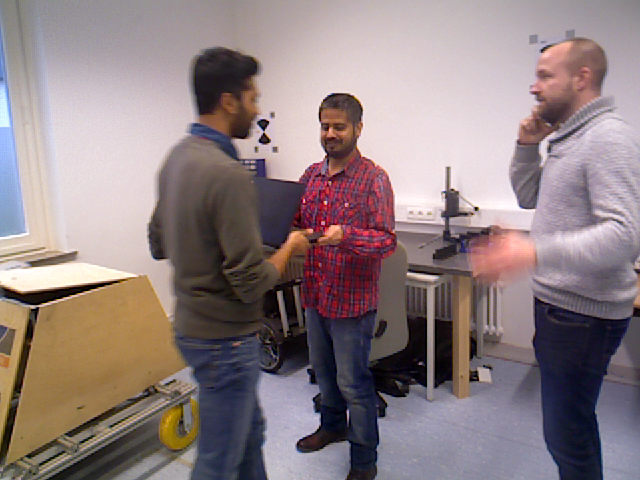}\label{fig:rgb_tomask}}\hspace{0.01pt}
  \subfloat[Residuals]{\includegraphics[width=0.48\columnwidth]{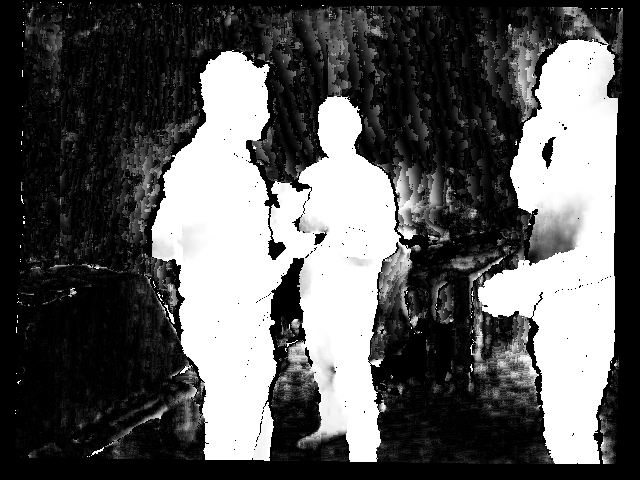}\label{fig:img_residuals}}\\
 \subfloat[Raw mask]{\includegraphics[width=0.48\columnwidth]{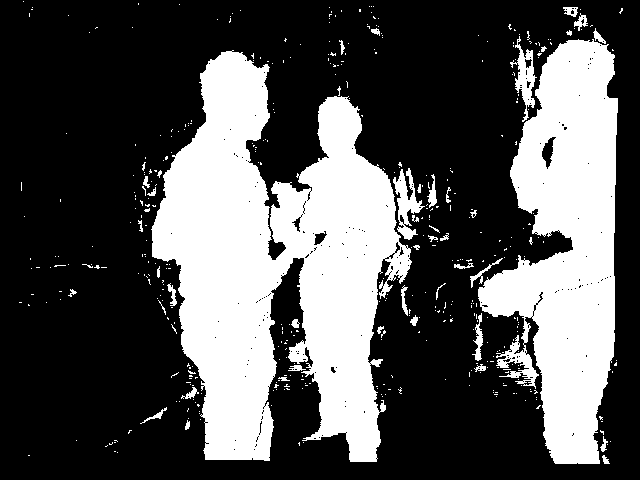}\label{fig:raw_mask}}\hspace{0.01pt}
 \subfloat[Final mask]{\includegraphics[width=0.48\columnwidth]{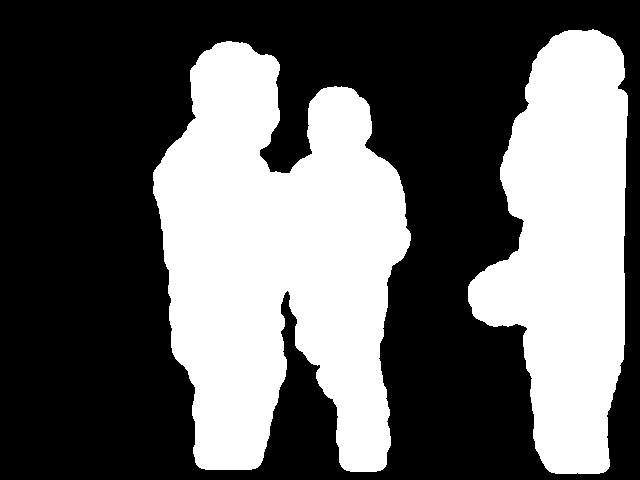}\label{fig:final_mask}}
  \caption{Steps of the mask creation. (a) Example RGB frame. (b) Residuals obtained from the registration. (c) Initial mask obtained from the residual. (d) Final refined mask after floodfill.}
\end{figure}

\begin{algorithm}[t]
\DontPrintSemicolon
\SetKw{Continue}{continue}
\KwIn{pixels of residual mask $\mathcal{M}_{\mathrm{R}}$}
\KwResult{Mask $\mathcal{M}_{\mathrm{D}}$}
  Let $\mathcal{Q}$ be a queue containing all the pixels to be masked.\; 
  Let $\mathcal{N}(\mathbf{p})$ be the set of neighbors of pixel $\mathbf{p}$\;
  Add all pixels from $\mathcal{M}_{\mathrm{R}}$ to queue $\mathcal{Q}$\;
  \While{$\mathcal{Q} \neq \emptyset$ }
  {
    Add all pixels inside $\mathcal{Q}$ to $\mathcal{M}_{\mathrm{D}}$\;
    \ForEach{$\mathbf{p} \in \mathcal{Q}$}
    {
      \ForEach{$\mathbf{n} \in \mathcal{N}(\mathbf{p})$}
      {
  \If{ $|| D(\mathbf{p}) - D(\mathbf{n}) || < \theta \cdot D(\mathbf{p})$} 
  {
    Add $\mathbf{n}$ to $\mathcal{Q}$ \textbf{if} $\mathbf{n} \notin \mathcal{M}_{\mathrm{D}}$\;
  }
      }
      Remove $\mathbf{p}$ from $\mathcal{Q}$\;
    }
    
  }
\caption{\textbf{floodfill} for generating $\mathcal{M}_{\mathrm{D}}$.}
\label{alg:floodfill}
\end{algorithm}
\begin{figure}[t]
\centering
\vspace{2mm}
\includegraphics[width=0.99\columnwidth]{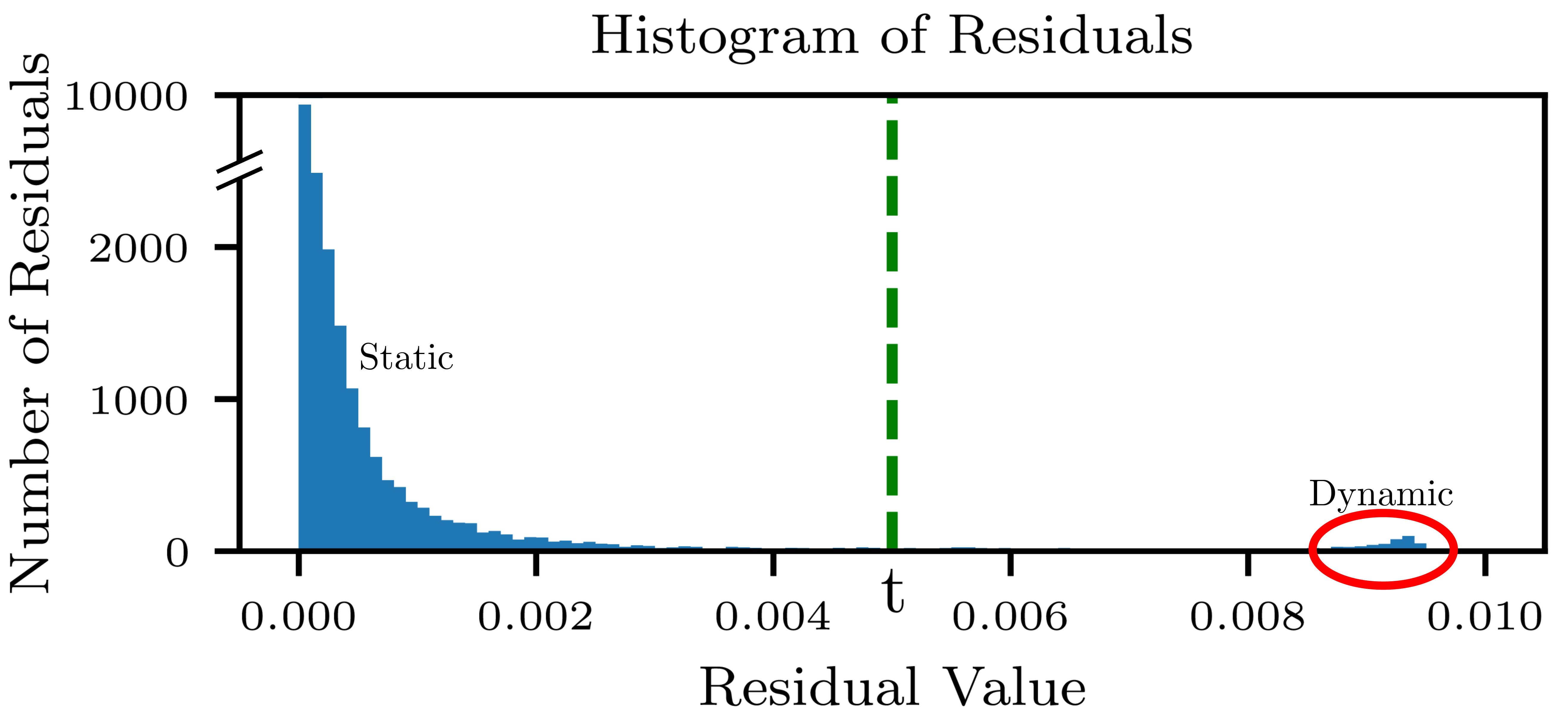}
\caption{Histogram of residuals obtained after the registration of an image containing dynamic elements. The red ellipse highlights the residuals resulting from the dynamic parts. By considering only pixels with residuals under the threshold $t$, we can identify the dynamic parts of the image.}
\label{fig:residuals_hist}
\end{figure}

To detect dynamic parts of the environment, we first perform an initial registration of the current RGB-D frame with respect to the model, as described in \secref{sec:pose_registration}. After registration, we compute for each pixel $\mathbf{p}_i$ its residual $r_i$ \wrt the model as defined in \eqref{eq:errordepth} and illustrated in \figref{fig:img_residuals}. We select a threshold:
\begin{eqnarray}
t &=& \gamma\tau^2,
\end{eqnarray}
where $\tau$ is the truncation distance used in our TSDF representation and $\gamma$ is a value between $0$ (everything is masked) and $1$ (nothing is masked). \figref{fig:residuals_hist} shows a histogram of the residuals obtained after the initial registration of one image. The figure shows how most of the residuals concentrate below a certain value, except the ones belonging to dynamic parts of the environment.
Every residual exceeding $t$ contributes to the creation of a binary mask, see \figref{fig:raw_mask}. Such threshold-based segmentation is often not perfect and may fail to capture the whole dynamic object. Since we have depth information available, we use the eroded mask $\mathcal{M}_{\mathrm{R}}$ to initialize a depth-aware flood fill algorithm, summarized in \algref{alg:floodfill}, similar to the region growing approaches used in~\cite{keller20133dv,ruenz2018ismar}. We add neighbors to a region as long as their depth does not differ more than a threshold $\theta$.
Finally, the mask is dilated again to cover eventual border pixels left out by the floodfill.
\figref{fig:final_mask} shows the resulting mask after all the processing steps. 
We then perform a second registration without masked pixels and we integrate the  RGB-D information ignoring the masked pixels into the model using the newly obtained pose. Note that, since our method performs a second registration step, the registration takes up to twice as long compared to the approach ignoring dynamics.

\subsection{Carving of Model and Free Space Management}
One weakness of TSDF approaches is that they cannot keep track of perceived free space. However, being aware of the previously measured free space is a way to reject dynamic objects, as opposed to detecting them from their motion. We follow a rather simple rule: a voxel that was found to be reliably empty can never contain a static object. Indeed, if a new range image points to a voxel being occupied, it can only be so because a dynamic object has entered it. Therefore, we can safely reject it. Doing so, we sidestep the notoriously difficult problem of tracking dynamic moving points, as done in~\cite{runz2017icra,ruenz2018ismar}. We mark as free every non-occluded voxel in the camera frustum outside the truncation region (up to a clipping plane). Note that this is particularly helpful in case of measurements of points that are too far from the sensor to be integrated into the TSDF. In our current implementation, we assign to free voxels an SDF value equal to the truncation distance $\tau$. Doing this for every voxel in the frustum is suboptimal in terms of memory consumption. Improvement by maintaining an octree-based representation of the free space is left as future work.

Another consideration regarding free space is that if a previously static object moves, its voxels must be removed from the map. This corresponds to the fact that if a voxel previously mapped as static is detected as empty subsequently, this voxel used to contain a dynamic object and should, therefore, be marked as free. This behavior is automatically achieved by the TSDF representation, \ie, we update the SDF stored in the voxel by performing a weighted average between the current value and the truncation distance $\tau$. Therefore, if a voxel is detected as free long enough, it will be reliably marked as free.

\subsection{Limitations and Handling Invalid Measurements}\label{sec:limitations}

\begin{figure}[t]
  \centering
  \subfloat[]{\includegraphics[width=0.45\columnwidth]{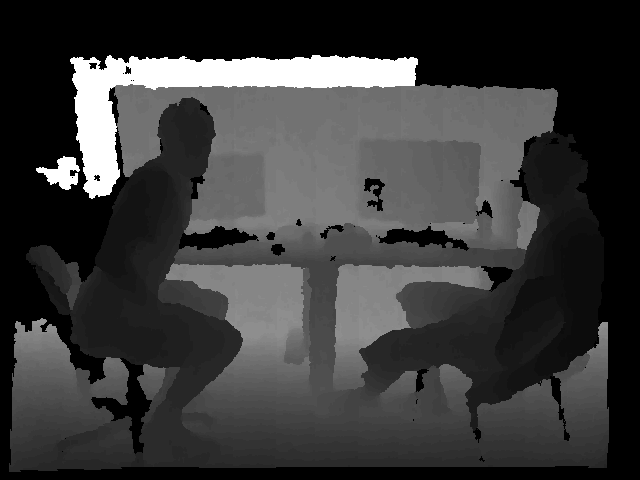}\label{fig:noisy_depth}}\hspace{0.2pt}
  \subfloat[]{\includegraphics[width=0.45\columnwidth]{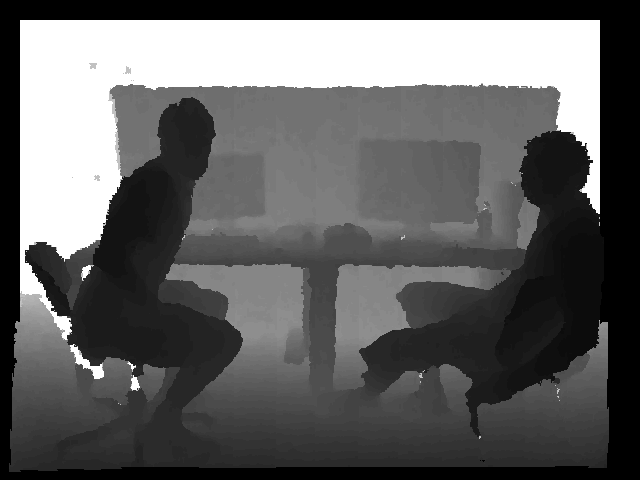}\label{fig:virtual_depth}}\\
  \caption{(a) Raw depth from the RGB-D sensor. (b) Our virtual refined depth from 10 adjacent frames. (\texttildelow0.3\,s delay)}
\end{figure}

Marking free voxels in the camera frustum is effective as long as we know from the sensor that those regions of space are empty. However, common commercial RGB-D cameras return depth images that contain invalid measurements, i.e., pixels with a value of zero, as exemplarily shown by black pixels in \figref{fig:noisy_depth}. These measurements are invalid either because they are out-of-range or because they are not measurable, \eg, because of scattering media, reflecting surfaces, \etc. For our approach to work in every possible case, it is necessary to distinguish between out-of-range and non-measurable values. Two methods to handle such cases are possible.

The first one is to make no distinction and not consider zero values, as it is commonly done in other RGB-D mapping approaches. 
In this case, our approach will work correctly when the whole scene is in the range of the depth sensor. 
A limitation of our approach is that in case there are out-of-range values, dynamic objects could be incorrectly added to the model, thus affecting its quality.

The second method is to ``correct'' non-measurable values if possible. 
To this end, we create a temporary model from $n$ consecutive frames and generate virtual depths from the registered poses.
Then, we fill in the original depth images by replacing every zero value with the corresponding value of the virtual depth. 
In this way, non-measurable values are usally reduced thanks to the multiple observation. 
The remaining values are assumed to come from out-of-range measurements of the camera and are replaced with a high fixed depth value. 
\figref{fig:virtual_depth} shows an example of a virtual depth image obtained with this technique. 
However, this solution assumes that nothing appears closer than the minimum range of the depth sensor. 
This assumption can be removed in case we add an additional sensor, \eg, a simple sonar on top of the RGB-D sensor, that detects whether there are objects too close to the camera. Moreover, a disadvantage of this method is that the actual model is then generated with a delay of $n$ frames, which might be suboptimal for some robotics applications.

In the following experiments, we employ the second option for modeling the sequences of the TUM RGB-D dataset, since the depth images contain out-of-range values. For the sequences of our dataset, we employ the first option, since the recorded depth is always within the valid range of the depth sensor.

\section{Experimental Evaluation}
\label{sec:exp}

\begin{table}[t]
\vspace{2mm}
\caption{Parameters of our approach used in all experiments.}
\centering
\begin{tabular}{C{4cm}C{2cm}}
\toprule
 Parameter & Value \\
\midrule
 Voxel size & $0.01$\,m \\
 Truncation distance $\tau$ & $0.1\,$m \\
 Weight $w_c$ & $0.025$ \\
 Floodfill threshold $\theta$ & $0.007$ \\
 Residual threshold weight $\gamma$ & $0.5$ \\
\bottomrule
\end{tabular}
\label{tab:parameters}
\end{table}

The main contribution of this work is a TSDF-based mapping approach that is able to operate in environments with the presence of highly dynamic elements by relying solely on geometric information, \ie, our approach is completely class agnostic and does not require tracking of objects.
Our experiments show the capabilities of our method and support our key claims, which are: (i) our approach is robust to dynamic elements in the environment and provides a camera tracking performance on par or better than  state-of-the-art dense SLAM approaches, and
(ii) provides a dense 3D model that contains only the static parts of the environment, which is more accurate than other state-of-the-art approaches when compared to the ground truth model.

We provide comparisons with StaticFusion (SF)~\cite{scona2018icra}, DynaSLAM (DS)~\cite{bescos2018ral} and MaskFusion (MF)~\cite{ruenz2018ismar}. As our approach does not rely on deep neural networks, we make a distinction in our comparison between the pure geometric approach of DynaSLAM (G) and the combined deep neural network+geometric approach (N+G).
We tested all approaches on the dynamic scenes of the TUM RGB-D dataset~\cite{sturm2012iros}, as well as on our dataset, designed to contain highly dynamic scenes. We obtained the reported results by using the open source implementations available for the different approaches, with the exception of MaskFusion, where we only report results from the paper~\cite{ruenz2018ismar}.

In all experiments, we used the default parameters provided by the open source implementations and the same holds for our approach, see \tabref{tab:parameters} for details. Our default parameters have been determined empirically by trial and error, but similar values gave comparable results.

In the presented tables, we separate the approaches that rely solely on geometric information, from approaches that rely also on neural networks. We highlight in bold the best result among the first category of approaches, as we focus mainly at class agnostic approaches.

\subsection{Performance on TUM RGB-D Dataset}

\begin{figure}[t]
  \centering
  \vspace{2mm}
 \includegraphics[width=0.5\columnwidth]{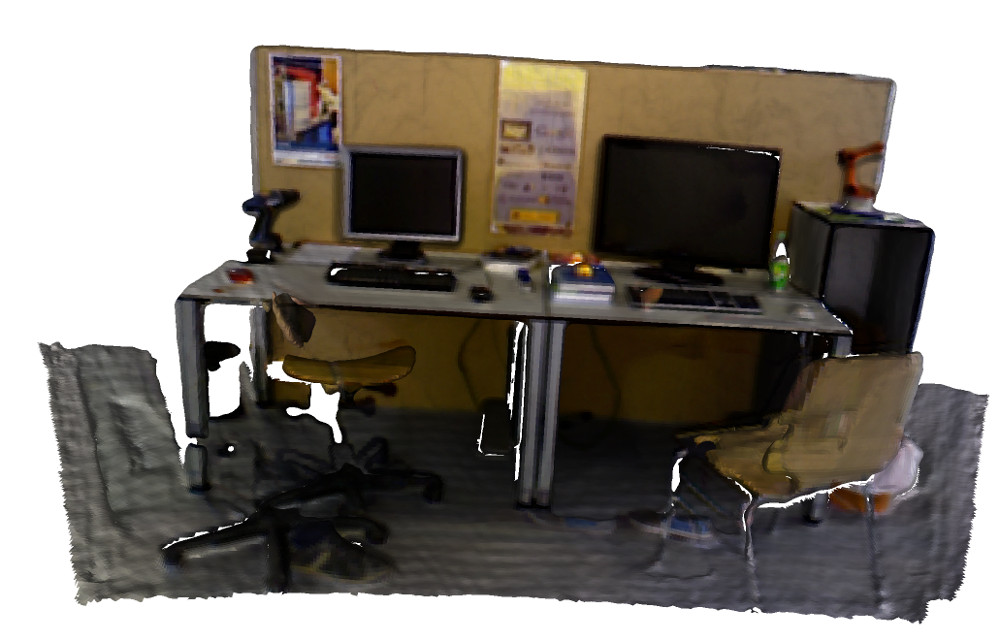}
  \caption{Final mesh obtained using our approach on the \emph{walking\_static} sequence, in which two people are continuously walking through the scene.}
  \label{fig:walking_static_model}
\end{figure}

\begin{table}[t]
\vspace{2mm}
\caption{Absolute Trajectory Error (RMS) [m] on dynamic scenes of TUM dataset.}
\centering
\begin{tabular}{cccc|cc}
\toprule
  & Ours & SF & DS (G) & DS (N+G) & MF \\
\midrule
 Dense approach & \cmark & \cmark & \xmark & \xmark & \cmark\\
 \midrule
 sitting\_static       & \textbf{0.009} & 0.014 & \textbf{0.009} & 0.007 & 0.021 \\
 sitting\_xyz          & 0.040 & 0.039 & \textbf{0.009} & 0.015          & 0.031 \\
 sitting\_halfsphere   & 0.110 & 0.041 & \textbf{0.017} & 0.028          & 0.052 \\
 walking\_static       & 0.017 & 0.015 & \textbf{0.014} & 0.007          & 0.035 \\
 walking\_xyz          & 0.099 & 0.093 & \textbf{0.085} & 0.017          & 0.104 \\
 walking\_halfsphere   & 0.104 & 0.681 & \textbf{0.084} & 0.026          & 0.106 \\
 \midrule
 Max                   & 0.110 & 0.681 & \textbf{0.085} & 0.028 & 0.106 \\
\bottomrule
\end{tabular}
\label{tab:tum_ate}
\end{table}

The first experiment shows the performance of our approach with the TUM RGB-D dataset~\cite{sturm2012iros}. 

Note that, since the depth information from these sequences contains out-of-range values, we used the approach described in~\secref{sec:limitations} to obtaine a refined depth, with the temporary model created from $n = 10$ consecutive frames, which corresponds to a model delayed approximately 0.3\,s.

\tabref{tab:tum_ate} shows the results of all considered approaches on six sequences of the TUM dataset. From this table, it is clear that DynaSLAM outperforms the other methods. However, DynaSLAM is a feature-based approach and, in contrast to the other approaches, does not provide a dense model. The three dense mapping approaches show in most of the cases similar results, except for the sequence  \emph{walking\_halfsphere}, where StaticFusion lost track due to the excess of dynamic elements at the beginning of the sequence, and the sequence \emph{sitting\_halfsphere}, where our approach shows worse performance. In terms of 3D reconstruction, our approach is always able to create a consistent mesh of the environment, see \figref{fig:walking_static_model} for an example.

\begin{figure}[t]
  \centering
  \vspace{2mm}
 \includegraphics[width=0.99\columnwidth]{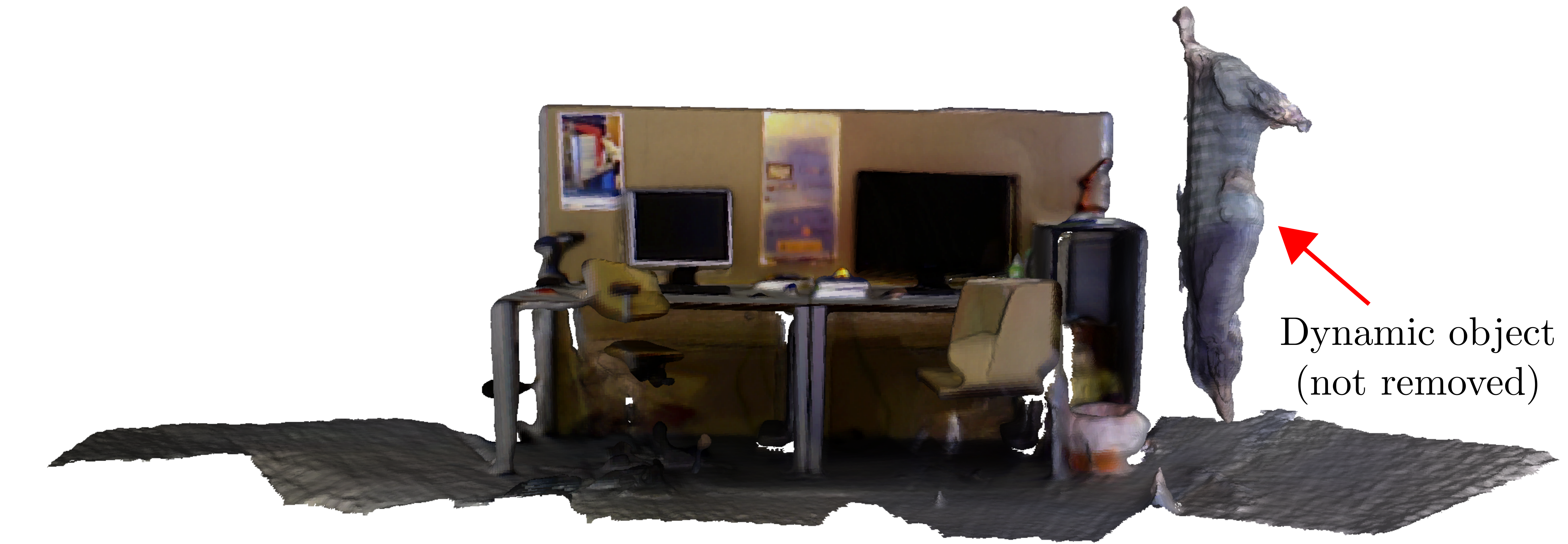}
  \caption{Final mesh obtained using our approach on the \emph{walking\_xyz} sequence. In this sequence, the camera follows the person on the right at the beginning and then never revisits that location. Therefore, the person is added in the model.}
  \label{fig:walking_xyz_model}
\end{figure}

\begin{figure}[t]
  \centering
  \vspace{-2mm}
 \includegraphics[width=0.95\columnwidth]{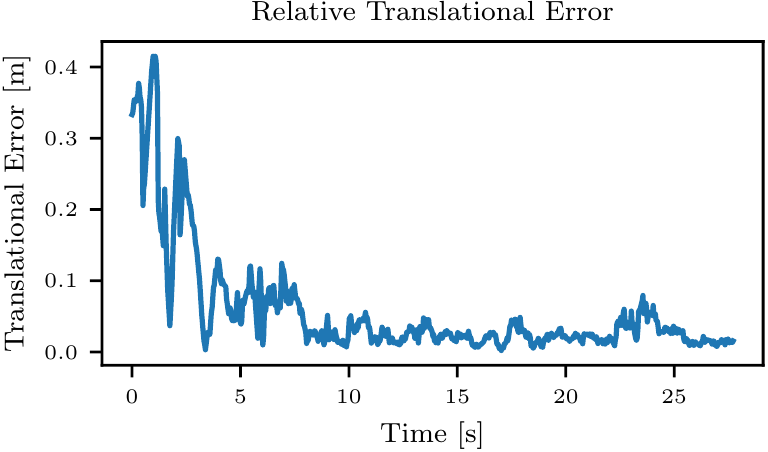}
  \caption{Relative translational error over time for the \emph{walking\_xyz} sequence. In this plot it is visible how the relative error is particularly high at the beginning of the sequence, when the camera is tracking the person. After the first four seconds, the error drops substantially.}
  \label{fig:rpe_walking_xyz}
\end{figure}

The only case where the model built by our approach shows artifacts is on the \emph{walking\_xyz} sequence, where a person remains in the model, see \figref{fig:walking_xyz_model}. This happens because the person is tracked by the cameraman at the beginning of the sequence and the location where the person stops is never revisited again. Therefore, the algorithm cannot know that the voxels in that location are actually free. This is confirmed by \figref{fig:rpe_walking_xyz}, which shows the relative position error versus the elapsed time. It is evident from the figure that in the first four seconds, \ie, when the camera tracks the person, the error is particularly high.
In sum, we are on par with state-of-the-art dense mapping approaches in terms of tracking, but we use a completely different technique based on TSDF instead of surfels. 

\subsection{Performance on Bonn RGB-D Dynamic Dataset}
\begin{figure}[t]
  \centering
  \subfloat{\includegraphics[width=0.32\columnwidth]{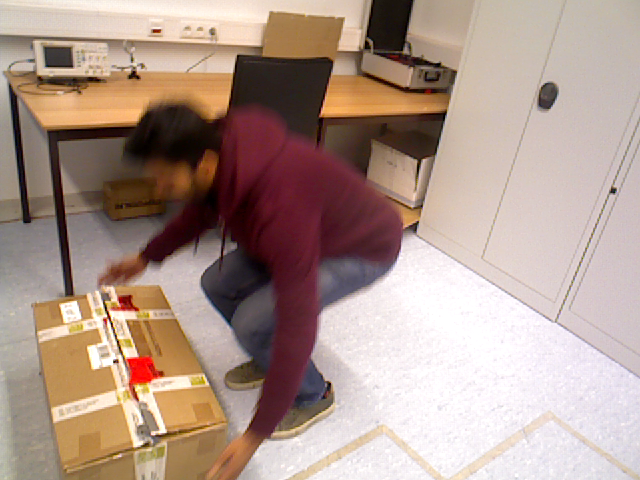}}\hspace{0.1pt}
  \subfloat{\includegraphics[width=0.32\columnwidth]{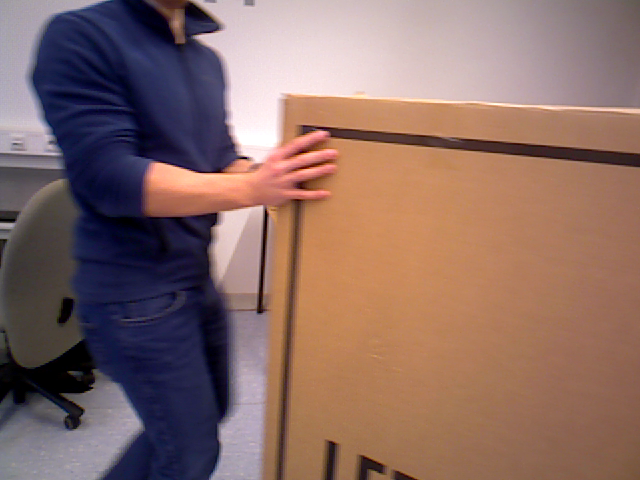}}\hspace{0.1pt}
 \subfloat{\includegraphics[width=0.32\columnwidth]{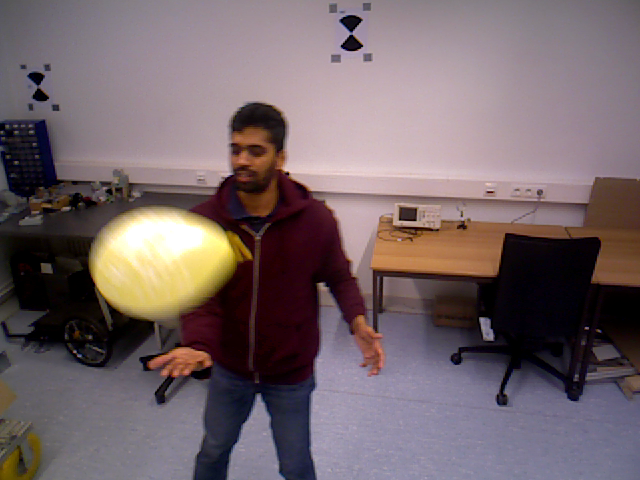}}
  \caption{Example RGB frames from our highly dynamic dataset.}
  \label{fig:dataset_example}
\end{figure}

\begin{table}[t]
\vspace{3mm}
\caption{Absolute Trajectory Error (RMS) [m] on our dataset. In this table, we shorten \emph{obstructing\_box} with \emph{o\_box} and \emph{nonobstructing\_box} with \emph{no\_box}.}
\centering
\begin{tabular}{cccc|c}
\toprule
 & Ours & SF & DS (G) & DS (N+G) \\
\midrule
 Dense approach & \cmark & \cmark & \xmark & \xmark\\
\midrule
 balloon              & 0.175          & 0.233           & \textbf{0.050}& 0.030 \\
 balloon2             & 0.254          & 0.293           & \textbf{0.142}& 0.029 \\
 balloon\_tracking    & 0.302          & 0.221           & \textbf{0.156}& 0.049 \\
 balloon\_tracking2   & 0.322          & 0.366           & \textbf{0.192}& 0.035 \\
 crowd                & \textbf{0.204} & 3.586           & 1.065         & 0.016 \\
 crowd2               & \textbf{0.155} & 0.215           & 1.217         & 0.031 \\
 crowd3               & \textbf{0.137} & 0.168           & 0.835         & 0.038 \\
 kidnapping\_box      & 0.148          & 0.336           & \textbf{0.026}& 0.029 \\
 kidnapping\_box2     & 0.161          & 0.263           & \textbf{0.033}& 0.035 \\
 moving\_no\_box      & \textbf{0.071} & 0.141           & 0.317         & 0.232 \\
 moving\_no\_box2     & 0.179          & 0.364           & \textbf{0.052}& 0.039 \\
 moving\_o\_box       & 0.343          & \textbf{0.331}  & 0.544         & 0.044 \\
 moving\_o\_box2      & 0.528          & \textbf{0.309}  & 0.589         & 0.263 \\
 person\_tracking     & \textbf{0.289} & 0.484           & 0.714         & 0.061 \\
 person\_tracking2    & \textbf{0.463} & 0.626           & 0.817         & 0.078 \\
 placing\_no\_box     & \textbf{0.106} & 0.125           & 0.645         & 0.575 \\
 placing\_no\_box2    & 0.141          & 0.177           & \textbf{0.027}& 0.021 \\
 placing\_no\_box3    & \textbf{0.174} & 0.256           & 0.327         & 0.058 \\
 placing\_o\_box      & 0.571          & 0.330           & \textbf{0.267}& 0.255 \\
 removing\_no\_box    & 0.041          & 0.136           & \textbf{0.016}& 0.016 \\
 removing\_no\_box2   & 0.111          & 0.129           & \textbf{0.022}& 0.021 \\
 removing\_o\_box     & \textbf{0.222} & 0.334           & 0.362         & 0.291 \\
 synchronous          & \textbf{0.441} & 0.446           & 0.977         & 0.015 \\
 synchronous2         & \textbf{0.022} & 0.027           & 0.887         & 0.009 \\
 \midrule
 Max                  & \textbf{0.571} & 3.586           & 1.217         & 0.575 \\
\bottomrule
\end{tabular}
\label{tab:bonn_ate}
\end{table}

The second set of experiments consists of the comparison between the algorithms on our dataset. Our dataset includes $24$ highly dynamic scenes, where people perform different tasks, such as manipulating boxes or playing with balloons, see~\figref{fig:dataset_example} for some example RGB frames. These tasks often obstruct the camera, creating particularly challenging situation for mapping approaches.
We recorded the dataset using an ASUS Xtion Pro LIVE sensor, combined with an Optitrack Prime 13 motion capture system for the ground truth trajectories. Additionally a Leica BLK360 terrestrial laser scanner was used to obtain a ground truth 3D pointcloud of the static environment.

\tabref{tab:bonn_ate} shows the performance of different approaches on our scenes. The variety of sequences show interesting phenomena. For example, on the scenes where the dynamic component is a uniformly colored balloon, DynaSLAM outperforms the dense approaches, because it cannot detect features on the balloon, which therefore does not affect the SLAM performance. Our approach performs best on scenes crowded with people, which are among the most challenging if no semantic segmentation algorithm is available. On sequences that involve the manipulation of boxes, the algorithms have mixed results, with our approach being better in about half of the cases and DynaSLAM being better on the other half. Note that DynaSLAM with the combined neural network and geometric approach performs the best in most cases. This is due to the heavy bias of having people in every sequence of our dataset, therefore the segmentation of people always helps the algorithm achieving better results. However, the worst performance of our approach is on par with the worst performance of DynaSLAM (N+G) and substantially better than the other geometric approaches, showing that our approach is more robust to failure. This is a desirable quality when deploying robotic systems. Finally, as discussed in the previous section, our approach is able to build a consistent model of the environment in most of the cases, see~\figref{fig:motivation} for an example of model from the scene \emph{crowd3}.

\subsection{Model Accuracy}

\begin{figure*}[t]
  \centering
  \subfloat[Ground truth point cloud]{\includegraphics[width=0.22\linewidth]{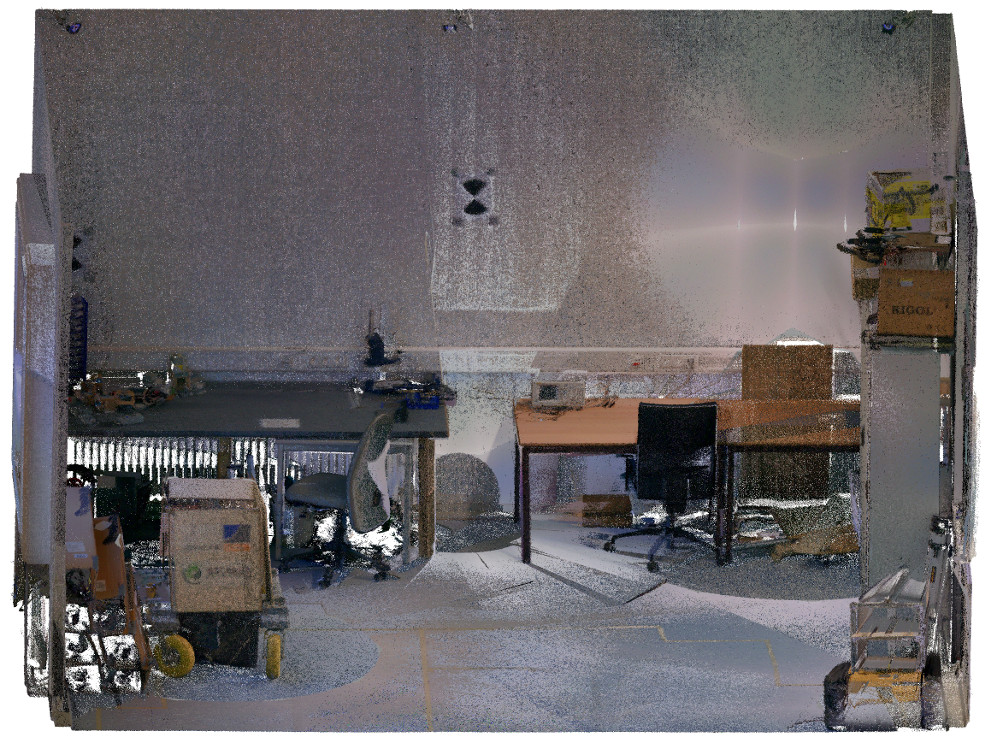}\label{fig:lab_groundtruth}}\hspace{0.02pt}
  \subfloat[ReFusion (our approach)]{\includegraphics[width=0.22\linewidth]{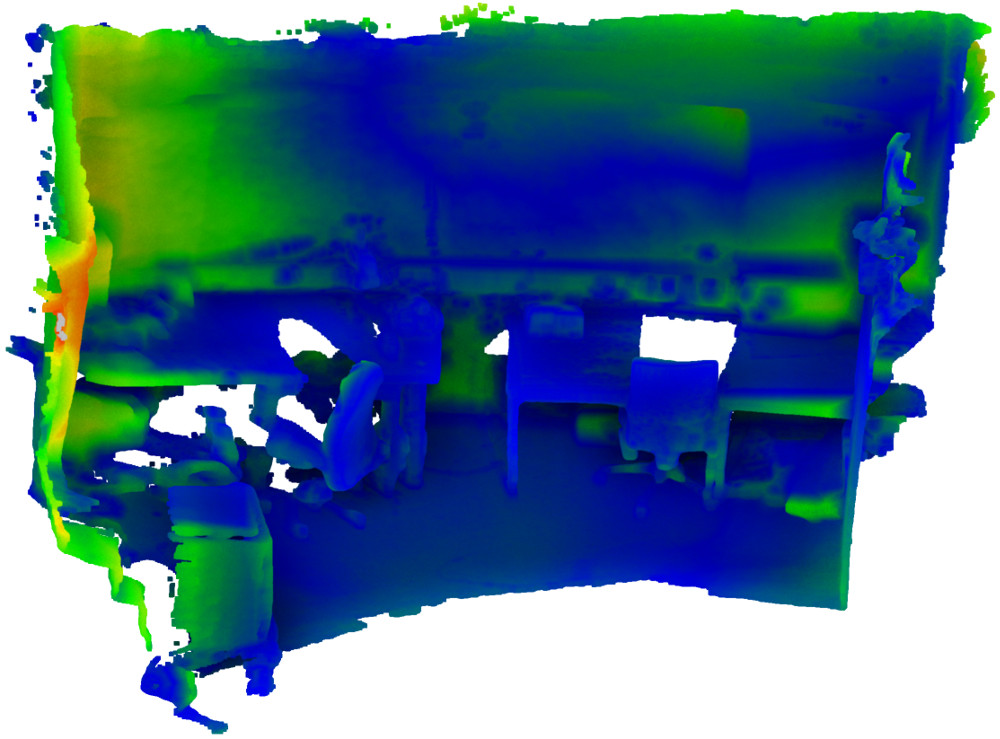}\label{fig:rf_crowd3_distance}}\hspace{0.02pt}
  \subfloat[StaticFusion]{\includegraphics[width=0.22\linewidth]{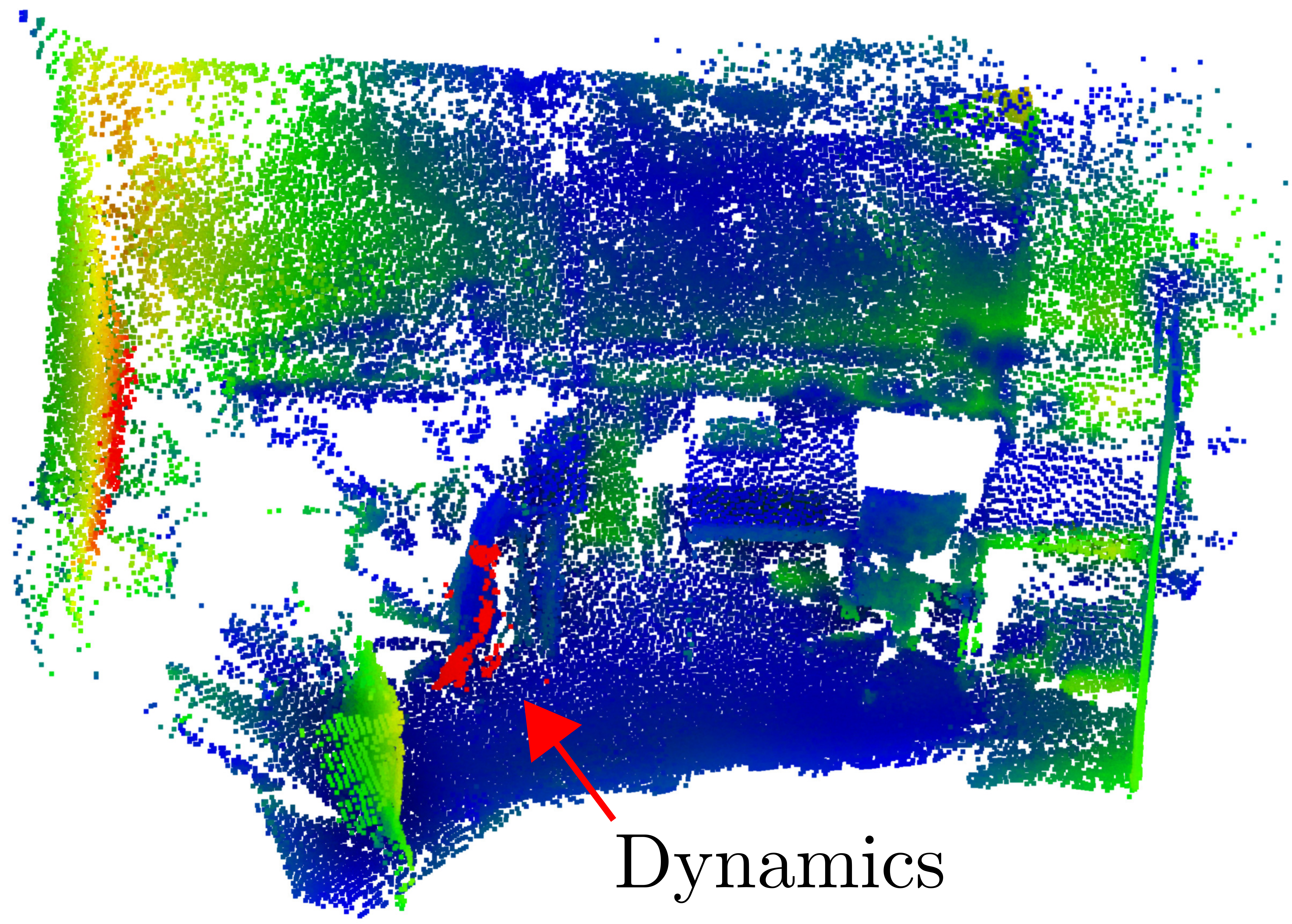}\label{fig:sf_crowd3_distance}}
  \subfloat{\includegraphics[width=0.04\linewidth]{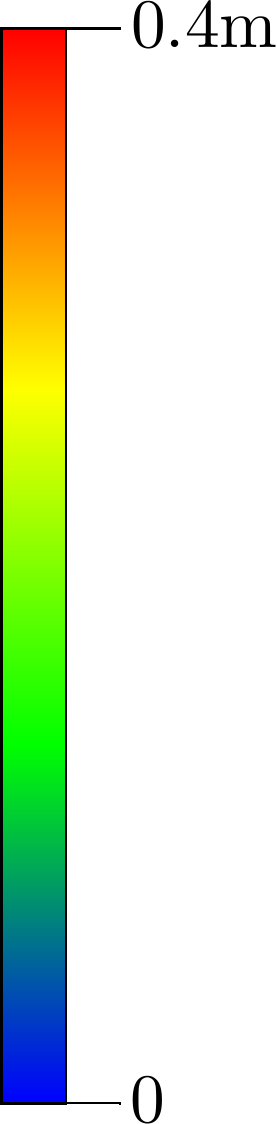}}
  \caption{Models from our approach and StaticFusion of the scene \emph{crowd3} compared against the ground truth. The points of the models are colored according to their distance from the ground truth. The arrow highlights the dynamic parts of the scene still present in the model from StaticFusion}
  \label{fig:model_accuracy}
\end{figure*}

\begin{figure}[t]
 \centering
 \includegraphics[width=0.9\columnwidth]{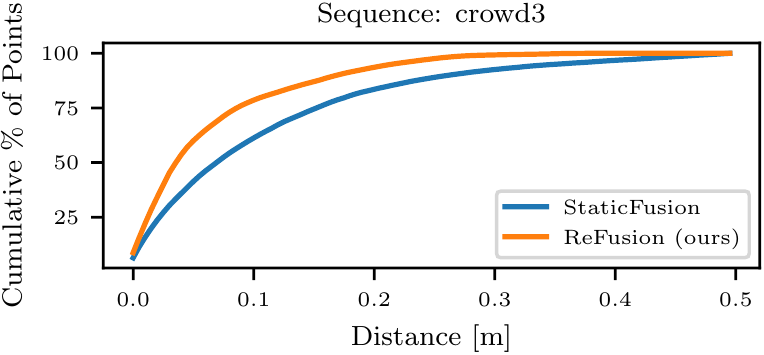}\\
 \vspace{-10pt}
 \includegraphics[width=0.9\columnwidth]{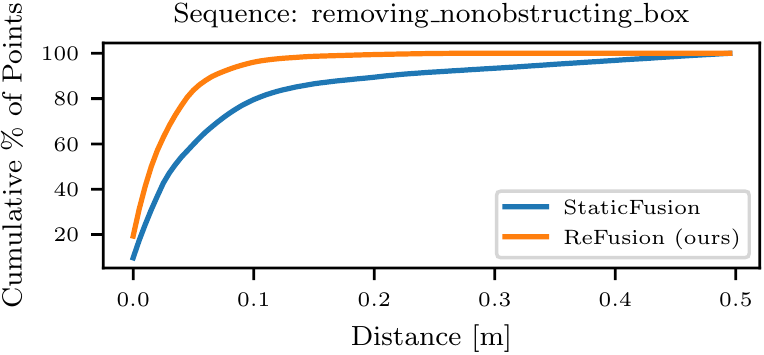}
 \caption{Plot of the cumulative percentage of points (y axis) at a specific distance from the ground truth 3D model (x axis). The higher the percentage of points towards a zero distance the better.}
 \label{fig:model_accuracy_plot}
\end{figure}

\begin{figure}[t]
  \centering
  \vspace{-3mm}
  \subfloat[]{\includegraphics[width=0.48\columnwidth]{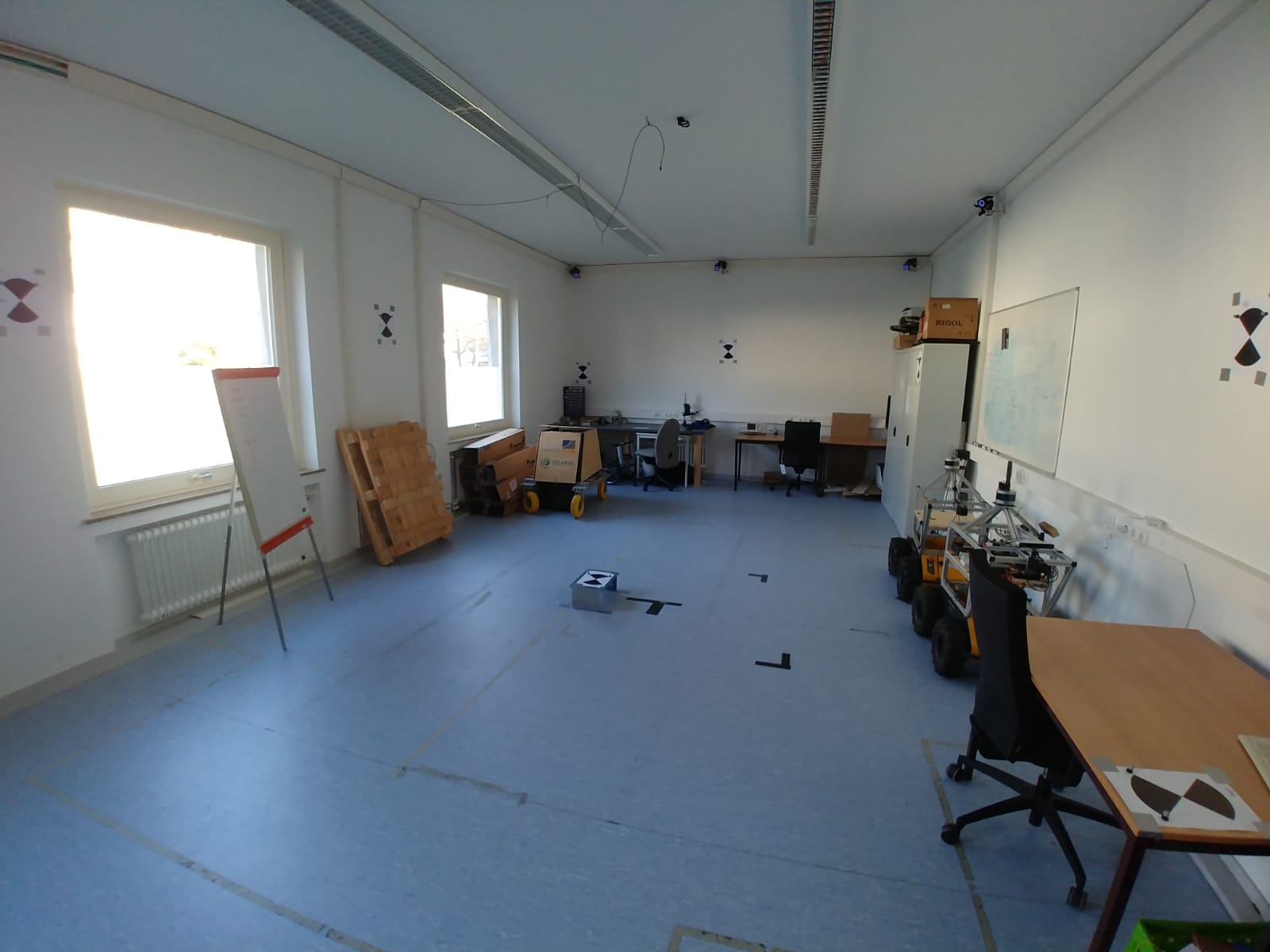}\label{fig:lab}}\hspace{0.01pt}
  \subfloat[]{\includegraphics[width=0.48\columnwidth]{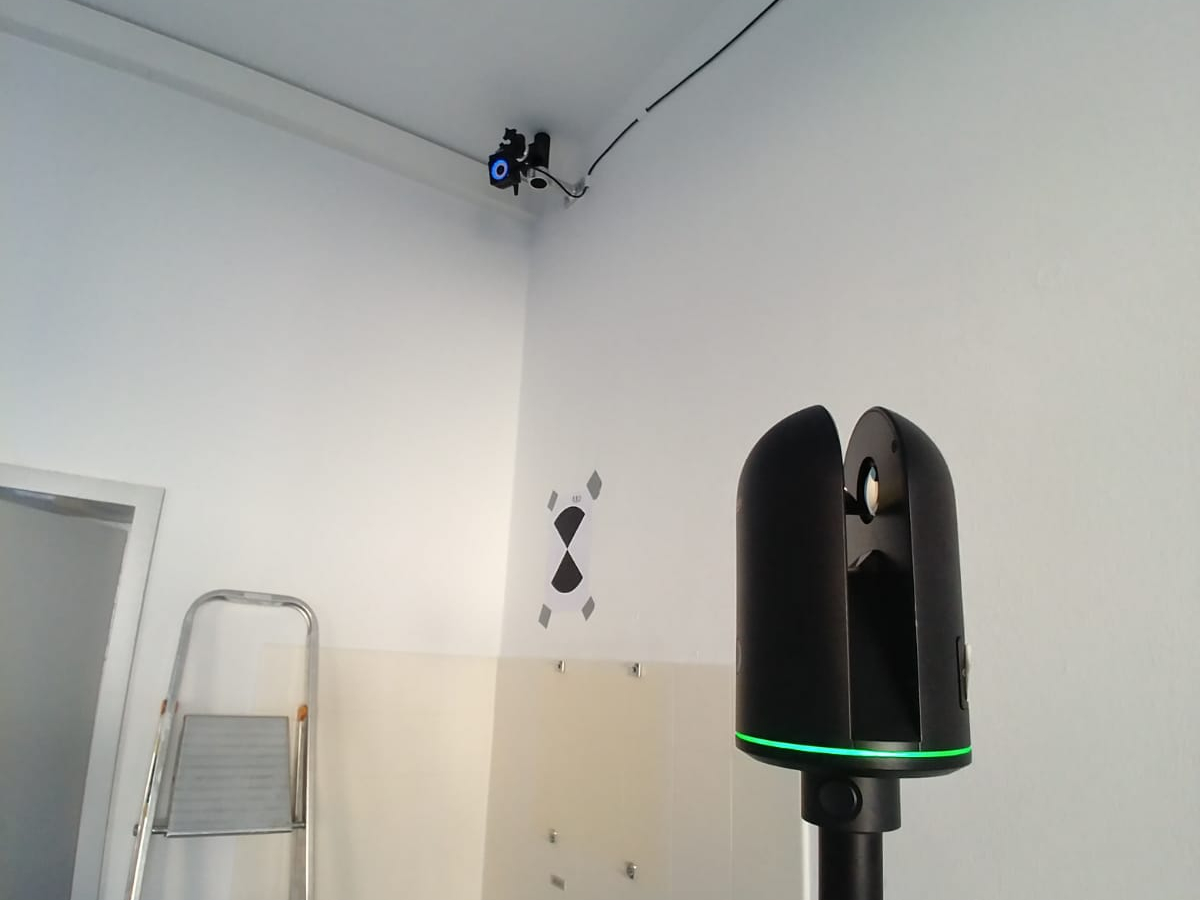}\label{fig:laser_scanner}}\\
  \vspace{-6pt}
  \subfloat[]{\includegraphics[width=0.48\columnwidth]{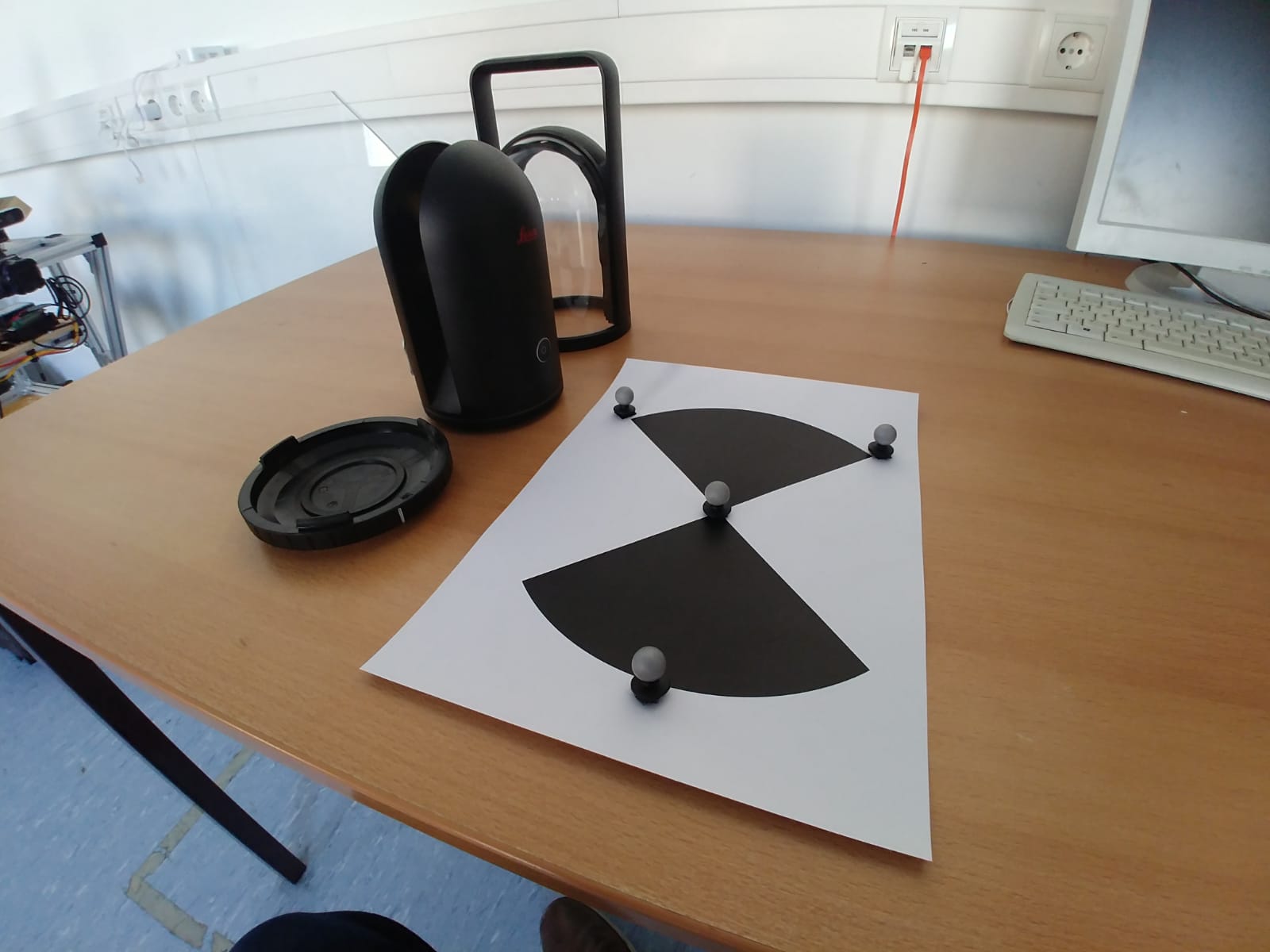}\label{fig:target}}\hspace{0.01pt}
 \subfloat[]{\includegraphics[width=0.48\columnwidth]{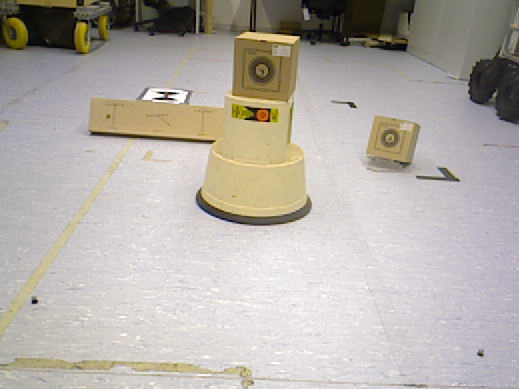}\label{fig:calib_setup}}
  \caption{(a) Our test environment. (b) Terrestrial laser scanner. (c) Tilt and turn target. (d) Calibration setup used to align the sensor's reference frame with the motion capture system's one.}
  
\end{figure}

The last set of experiments shows that our approach provides an accurate, dense 3D model that contains only the static parts of the environment. 

To perform such experiments, we first built a high resolution point cloud of the static part of our test environment (\figref{fig:lab}) using a professional terrestrial laser scanner, the Leica BLK360 (\figref{fig:laser_scanner}). We then aligned the point cloud to our motion capture system's reference frame using tilt and turn targets (\figref{fig:target}) that we located with both the laser scanner and the motion capture system. \figref{fig:lab_groundtruth} shows a section of our ground truth point cloud.

To align the model created by the algorithms to our ground truth, we transformed it from the reference frame of the RGB-D sensor, to the reference frame of the motion capture system. We aligned the two frames using the calibration setup shown in \figref{fig:calib_setup}, where the markers positioned in the environment were known in both reference frames.

We compare the models built by our algorithm and by StaticFusion~\cite{scona2018icra} for the sequences \emph{crowd3} and \emph{removing\_nonobstructing\_box} \wrt the ground truth. For each point of the evaluated model, we measure its distance from the ground truth. 

For a qualitative impression, \figref{fig:model_accuracy} shows the two models of the scene \emph{crowd3} where the points have been colored according to their distance to the closest point in the ground truth model. In \figref{fig:sf_crowd3_distance}, one can see that some dynamic elements are still present in the final model, represented by the red points highlighted by the arrow. In contrast, the model from our approach does not show such artifacts caused by dynamic objects. 

For a quantitative evaluation, \figref{fig:model_accuracy_plot} shows the cumulative percentage of points at a certain distance from the ground truth for the models of the two considered sequences. The plots show in both cases that the reconstructed model by our approach is more accurate.

In summary, our evaluation shows that our method is able to robustly track an RGB-D sensor in highly dynamic environments. At the same time, it provides a consistent and accurate model of the static part of the environment.

\section{Dataset and Source Code}
Our Bonn RGB-D dynamic dataset is available at the URL: \href{http://www.ipb.uni-bonn.de/data/rgbd-dynamic-dataset/}{http://www.ipb.uni-bonn.de/data/rgbd-dynamic-dataset}. The source code of our approach is available at the URL: \href{https://github.com/PRBonn/refusion}{https://github.com/PRBonn/refusion}.

\section{Conclusion}
\label{sec:conclusion}

We presented ReFusion: a TSDF-based mapping approach able to track the pose of the camera in dynamic environments and build a consistent 3D model of the static world.
Our approach tracks the sensor by exploiting directly the TSDF information and
the color information encoded in voxel blocks that are only allocated when needed.
Our method filters
dynamics using an algorithm based on the residuals from the registration and the representation of free space. We
evaluated our approach on the popular TUM RGB-D dataset, as well as on our Bonn RGB-D dynamic dataset,
and provided comparisons to other state-of-the-art techniques. Our experiments
show that our approach leads to an improved pose estimation in the presence of 
dynamic elements in the environment, compared to other state-of-the-art dense SLAM approaches.
Finally, we publicly release our own dataset, as well as our open-source implementation of the approach.

\section*{Acknowledgments}
We thank Raluca Scona for the assistance in running StaticFusion and Berta Bescos for the assistance in running DynaSLAM. Furthermore, we thank Jannik Jan{\ss}en, Xieyuanli Chen, Nived Chebrolu, Igor Bogoslavskyi, Julio Pastrana, Peeyush Kumar, Lorenzo Nardi, and Olga Vysotska for supporting the Bonn RGB-D dynamic dataset acquisition.

\bibliographystyle{plain}

\bibliography{palazzolo2019iros}

\end{document}